\documentclass[]{spie}  

\usepackage{amsmath,amsfonts,amssymb, bm}
\usepackage{graphicx}
\usepackage[colorlinks=true, allcolors=blue]{hyperref}

\usepackage{soul}

\usepackage[table,xcdraw]{xcolor}
\usepackage{color}
\usepackage{tabularray}
\usepackage{subfigure}

\definecolor{Alto}{rgb}{0.93,0.93,0.93}
\definecolor{JapaneseLaurel}{rgb}{0,0.6,0}
\definecolor{Roti}{rgb}{0.776,0.69,0.266}
\definecolor{Broom}{rgb}{0.984,1,0.074}
\definecolor{Lime}{rgb}{0.713,1,0.019}
\definecolor{SpringGreen}{rgb}{0.152,1,0.529}
\definecolor{Crail}{rgb}{0.76,0.309,0.266}
\definecolor{SilverChalice}{rgb}{0.647,0.647,0.647}
\definecolor{Aquamarine}{rgb}{0.411,1,0.972}
\definecolor{Milan}{rgb}{0.976,1,0.643}
\definecolor{Blue}{rgb}{0.109,0.05,1}

\title{A deep multiple instance learning approach based on coarse labels for high-resolution land-cover mapping}

\author{Gianmarco Perantoni}
\author{Lorenzo Bruzzone}
\affil{Department of Information Engineering and Computer Science, University of Trento, Via Sommarive, 5 I-38123, Trento, Italy}

\authorinfo{Proc. SPIE 12733, Image Sign. Proc. Remote Sens. XXIX, 127330H, Preprint, Full version: \href{https://doi.org/10.1117/12.2679464}{10.1117/12.2679464}\\Further author information:\\Gianmarco Perantoni: E-mail: gianmarco.perantoni@unitn.it\\  Lorenzo Bruzzone: E-mail: lorenzo.bruzzone@unitn.it}

\begin{document} 
\maketitle

\begin{abstract}
The quantity and the quality of the training labels are central problems in high-resolution land-cover mapping with machine-learning-based solutions. In this context, weak labels can be gathered in large quantities by leveraging on existing low-resolution or obsolete products. In this paper, we address the problem of training land-cover classifiers using high-resolution imagery (e.g., Sentinel-2) and weak low-resolution reference data (e.g., MODIS -derived land-cover maps). Inspired by recent works in Deep Multiple Instance Learning (DMIL), we propose a method that trains pixel-level multi-class classifiers and predicts low-resolution labels (i.e., patch-level classification), where the actual high-resolution labels are learned implicitly without direct supervision. This is achieved with flexible pooling layers that are able to link the semantics of the pixels in the high-resolution imagery to the low-resolution reference labels. Then, the Multiple Instance Learning (MIL) problem is re-framed in a multi-class and in a multi-label setting. In the former, the low-resolution annotation represents the majority of the pixels in the patch. In the latter, the annotation only provides us information on the presence of one of the land-cover classes in the patch and thus multiple labels can be considered valid for a patch at a time, whereas the low-resolution labels provide us only one label. Therefore, the classifier is trained with a Positive-Unlabeled Learning (PUL) strategy. Experimental results on the 2020 IEEE GRSS Data Fusion Contest dataset show the effectiveness of the proposed framework compared to standard training strategies. 
\end{abstract}


\section{INTRODUCTION}
\label{sec:intro}  
Nowadays, advancements in sensing technologies have significantly enhanced the spatial, spectral, and radiometric resolutions of optical and radar sensors aboard Earth-orbiting satellites. This progress, along with the establishment of satellite constellations, has opened up new possibilities for acquiring vast amounts of content-rich data with shorter revisit times. Notably, various satellite programs, such as the European Space Agency's (ESA) Copernicus programme featuring Sentinel satellites, have been designed to make this data readily accessible to both researchers and industries. Open and free access to high-resolution satellite data has revolutionized remote sensing applications, including global land-cover mapping. To process and analyse this wealth of data, the development of automatic image analysis techniques becomes imperative. Traditional machine learning methods face limitations when dealing with the complexity and variability of high-resolution satellite data on a large scale, resulting in poor generalization. Therefore, there has been a paradigm shift towards deep learning solutions, specifically deep neural networks (DNNs). DNNs have demonstrated remarkable performance in various domains due to their ability to automatically learn complex feature representations from the data \cite{8113128}.
\par
In remote sensing image analysis, deep learning has found extensive use, particularly with optical data like multi-spectral and hyper-spectral images. However, DNNs do require a large training dataset, which can be particularly challenging for classification tasks such as land-cover mapping due to the need for reference annotations. Furthermore, transitioning from low-resolution to high-resolution mapping introduces complexities in the data, making the availability of accurate reference labels crucial. This poses a challenge for DNNs to generalize correctly to large-scale applications, as the training set may not adequately represent the spectral characteristics of land covers.
\par
The quantity and the quality of the training labels are therefore central problems in high-resolution mapping. On the one hand, in computer vision, the problem of collecting sufficiently large datasets is usually addressed with crowdsourcing, which in turn produces noisy labels. Then, ad-hoc strategies (weakly supervised methods \cite{zhou2018brief}) are used to deal with the noisy labels. On the other hand, in remote sensing the collection of highly reliable training databases is a time-consuming task that requires expertise and thus cannot rely on crowdsourcing. Nonetheless, large quantities of noisy labels can be gathered by leveraging on existing low-resolution or obsolete products \cite{bruzzone2019multisource}, and thus weakly supervised strategies can be adopted \cite{cordeiro2020survey}. 
Among them, weak supervision with coarse-grained labels is usually referred to as Multiple Instance Learning (MIL)\cite{foulds2010review}. Standard MIL techniques are defined over binary classification tasks. The general MIL assumption states that training labels are given to bags (i.e., groups) of training instances and each bag is labelled as positive only if one or more instances in the bag are positive, meaning that the knowledge of the actual positive instances is missing and has to be inferred. Hence, the predictions of an instance-level classifier are aggregated by means of a pooling function (e.g., mean or max pooling) to generate the bag-level predictions, which are then compared with the given bag label. MIL was originally designed for shallow machine learning models, and only recently its combination with neural network gained interest. Multiple Instance Neural Networks (MINNs) have been reviewed by Wang \textit{et al.}\cite{wang2018revisiting}, who analysed and proposed architectures with different standard MIL pooling layers [i.e., max, mean and log-sum-exp (LSE)]. Attention-based Deep MIL (ADMIL) \cite{ilse2018attention} defines a pooling operation based on the attention mechanism, which both allows to dynamically select the relevant instances in a bag and is differentiable, making it appealing to use along with DNNs. Loss-based Attention for Deep MIL \cite{shi2020loss} consists in a strategy where the instance-level loss is exploited to compute the attention used in the pooling operation, which is then combined with a consistency loss that performs self-ensembling as a form of regularization. Examples of real-case applications of MIL are medical diagnosis and semantic segmentation. Semantic segmentation, in particular, can be seen as a MIL problem when the labels are given at image level. In the last few years, weakly supervised semantic segmentation (WSSS) gained interest in computer vision literature \cite{chan2021comprehensive}. In remote sensing literature, only recently WSSS gained interest. Some papers \cite{chan2021comprehensive, wang2020weakly} focused on CAM-based methods from the computer vision literature. Others considered the similarity of feature vectors of the pixels to group them before the actual segmentation \cite{nivaggioli2019weakly, ahlswede2022weakly}.
\par
The main and easiest way to retrieve sources of labels for land-cover mapping are the already available digital maps.  When using the available maps as reference during the training of a DNN, inaccuracy, land-cover changes and spatial resolution are the main sources of errors that can mislead the learning process. However, it is possible to address these issues by using robust training approaches. In this paper, we address the challenge posed using coarse resolution maps for higher resolution mapping using deep MIL (DMIL) based approaches. Up to our knowledge, no prior work tried to exploit a full MIL formulation to train DNN classifiers for land-cover mapping with low-resolution reference maps. Inspired by ADMIL \cite{ilse2018attention}, we propose a method that trains a pixel-level multi-class DMIL DNN to predict low-resolution labels (i.e., patch-level labels) while also learning the actual high-resolution labels without direct supervision. This is achieved under two possible alternative assumptions: 
i) \textit{multi-class}, a low-resolution label provides us the land-cover class for the majority of the pixels in a patch (i.e., which land cover is dominant),  and ii) \textit{multi-label}, a low-resolution label provides us a land-cover class for any of the pixels in the patch (i.e., which land cover is present, however many land covers are present).
On the one hand, the first one is a strong assumption, strictly related to the low-resolution map accuracy. However, it makes it possible to solve the MIL classification problem in a multi-class single-label setting. On the other hand, the second assumption is weaker and thus easier to satisfy. However, the problem becomes a multi-label classification problem where only some positive labels are given, as multiple land covers can be present in a given image patch and the low-resolution labels provide us only one label for each patch. Therefore, in order to train a DNN under this assumption, a Positive-Unlabeled Learning (PUL) \cite{niu2016theoretical} approach is necessary. PUL refers to the binary classification task where only some positive labels are given for training, and the remaining training data is unlabelled. Theoretical analyses \cite{du2014analysis} showed that given the knowledge of the prior distribution of the positive and negative classes and the use of a symmetric loss function, an unbiased estimator of the risk can be formulated. However, the empirical risk on training data can be negative, causing overfitting. This behaviour becomes more relevant for flexible models, such as DNNs. To solve this problem, a non-negative risk estimator (nnPU) \cite{kiryo2017positive} was developed that explicitly constrains the risk to be non-negative and that can be used with SGD-like stochastic optimization algorithms.
\par
In this paper, we propose the combination of MIL and PUL to learn high-resolution land-cover deep classifiers using low-resolution land-cover maps. This combination was used in the literature for defining a convex classification method for Positive-Unlabeled Multiple Instance Learning (PU-MIL) based on set kernel classifiers. However, up to our knowledge, no PU-MIL strategies have ever been proposed to tackle the training of a multi-class DNN-based land-cover classifier using coarse-resolution land-cover maps. The contributions of this paper can be summarized as follows: i) a PU-MIL method is proposed, combining nnPU with DMIL; ii) a loss function is presented combining the multi-class and multi-label assumptions, and iii) we perform a comparison between MIL methods and a standard training strategy that neglects the low-resolution nature of the coarse-grained labels, using them as if they were fine-grained. The experimental results were generated by mapping the land cover at $10m$ resolution using Sentinel-2 multispectral images and MODIS-derived land-cover map at $500m$ resolution as training labels. Then, the resulting high-resolution land-cover maps were validated against the $10m$ resolution reference map of the 2020 IEEE GRSS Data Fusion Contest (DFC2020).

\section{PROPOSED PU-DMIL APPROACH}
\label{sec:method}  
This section first presents the problem definition, along with the key insights that allow us to formulate a PU-MIL problem in the land-cover mapping scenario. Then, the details of the proposed architectures used in the experiments are described.
\subsection{Multiple-Instance Learning Models}
\label{sec:mil_method}  
In the standard binary supervised learning scenario, the objective is to find a model that correctly predicts the target variable $y \in \{ -1,+1 \}$ for an instance $\mathbf{x} \in \mathbb{R}^n$. To achieve this, a training set composed of tuples $(\mathbf{x}_i, y_i)$ is used to train the model, where each instance $\mathbf{x}$ is associated to a label $y$. However, in the MIL framework, we have instead labels $Y$ associated to bags of instances $X= \{ \mathbf{x}_1, ..., \mathbf{x}_K \}$. We assume that each of the $K$ instances $\mathbf{x}_j$ in the bag is associated to an unknown label $y_j$, which remain unknown at training time. However, bag labels provide coarse-grained information about the instance-level labels. Indeed, the common assumption of the MIL problem can be re-written in the following form:
\begin{equation} \label{eq:1}
    Y=\max_{j=1,...,K}\left\{y_j\right\}.
\end{equation}
This formulation is well suited for the training of high-resolution (HR) land-cover classifier with low-resolution (LR) labels. Indeed, the minimum mapping unit (MMU) of the reference map covers an area that is larger than the pixel size of the HR satellite image used to classify the land cover. Thus, the MMU of the reference map contains different HR satellite observations, and the actual correspondence of the LR land-cover label with these observations is unknown. However, we can assume that at least one of these observations actually belongs to the provided LR label.  Therefore, the MIL formulation allows us to take into consideration this uncertainty during the training process to improve the performance compared to a method that simply considers the LR label as the correct label for all the observations within its MMU. This would mean that many of these associations are incorrect, misleading the training process. Here, we consider satellite images patches as bags, each associated to a single LR land-cover label.
\par
An important observation at the basis of the proposed method is that an accurate land-cover classifier is also an accurate bag classifier. However, the contrary is not necessarily the case, as different HR classifications can be mapped to the same LR label due to the assumption of Equation (\ref{eq:1}) being permutation invariant. Nonetheless, if the DNN used is expressive enough, we can assume that feature vectors of observations of the same class will be close (i.e., cluster assumption), making it more likely that the learned HR classification with MIL will be close to the ground truth. Moreover, additional regularization can be included to further alleviate the effect of the ill-posedness of the MIL problem by exploiting the spatial correlation of pixels in the high-resolution satellite image with Convolutional Neural Networks (CNNs).
\par
\begin{figure} [t]
   \begin{center}
   \begin{tabular}{c} 
   \includegraphics[height=9cm]{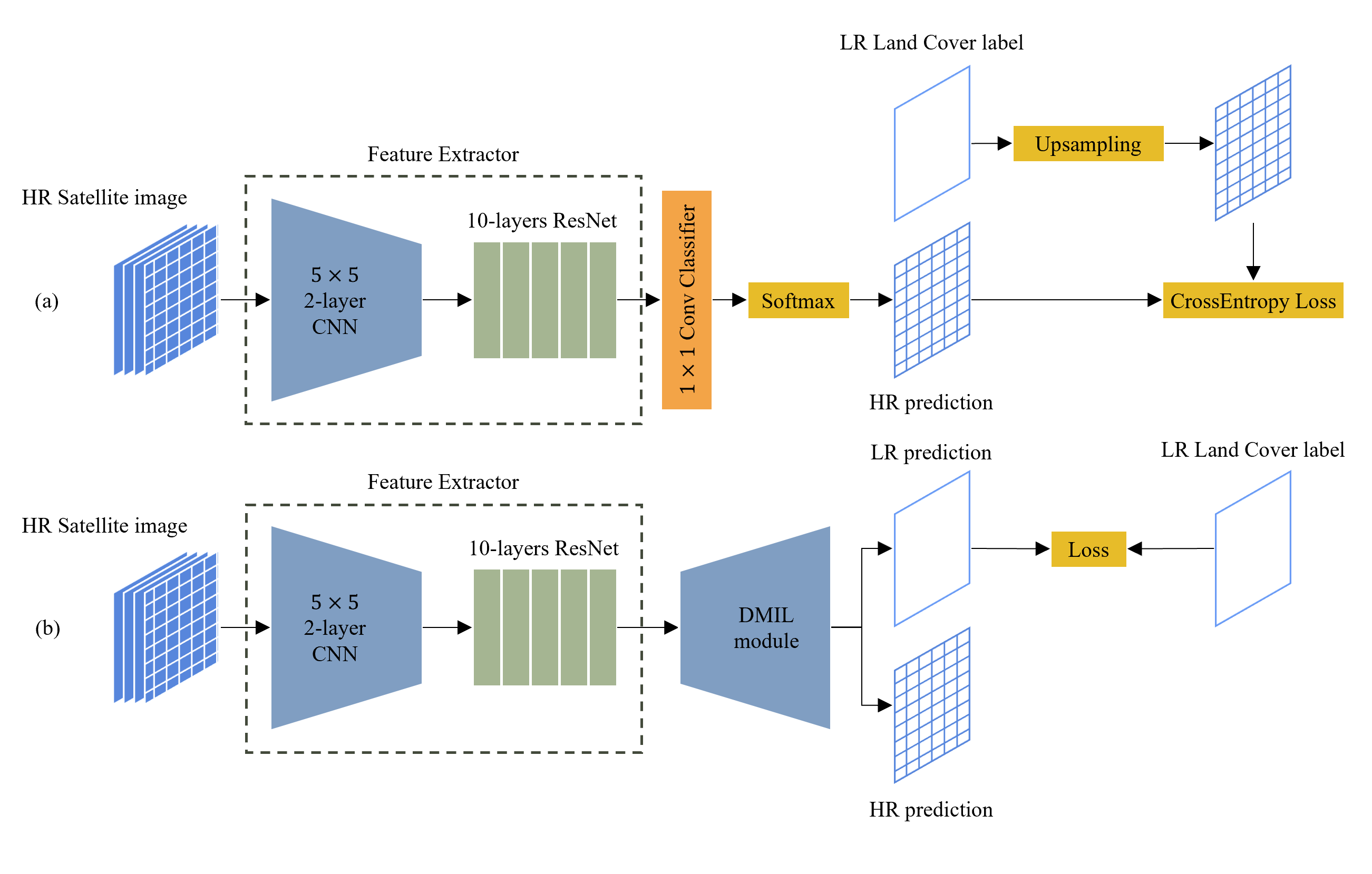}
   \end{tabular}
   \end{center}
   \caption[method] 
   { \label{fig:method} 
Architecture of the baseline and the proposed methods. (a) Baseline architecture, where the LR label is up-sampled and used as it was an HR label. (b) Proposed architecture, with a DMIL module that produces both high- and low- resolution predictions of the land cover, allowing to compute the loss directly at the resolution of the training label.}
   \end{figure}

One should note that too large receptive fields will reduce the actual perceived resolution of the generated land-cover map, as small scale details will likely be smoothed out. Thus, a trade-off is necessary, especially when the scale of the target resolution is such that objects can be as small as the pixel size (e.g., Sentinel-2 $10m$ resolution). Fig. \ref{fig:method} shows the proposed architecture used in this paper, where the feature extraction part of the DNN is constructed by combining a simple 2-layer CNN with a combined receptive field of $5 \times 5$ pixels and a 10-layer ResNet, where all the convolutions have $1 \times 1$ spatial kernels. Fig. \ref{fig:method}(a) shows the standard architecture used as baseline, where the LR map used as reference during training is upscaled to the resolution of the HR satellite image by means of nearest-neighbour interpolation. Fig. \ref{fig:method}(b) shows instead the proposed approach, where the DMIL module is used to generate both the high- and low- resolution land-cover predictions.
\par
Learning a model under the MIL assumptions requires optimizing an objective on the maximum over instances in a bag, which is problematic as DNNs optimization is typically gradient-based. For these reasons, different approaches have been proposed in the literature \cite{ilse2018attention}. The main component of a MIL learning strategy is the MIL pooling operation, which aggregate the contributions of each element in the bag in a way that allows to identify possible positive instances. Given a pooling operator, we can identify two main approaches that differ in how this contribution is defined: i) the \textit{instance-level} approach computes the class score for each instance before the pooling operation, where ii) the \textit{embedding-level} approach performs the pooling operation on the feature representation of the instances in the bag, producing a bag feature representation that will be then classified to produce the bag-level prediction. Usually, the pooling operations can be used independently of the approach chosen, however, each approach has its advantages and disadvantages. For convenience, let us consider the embedding approach throughout the paper. Let $\mathbf{h}_j \in \mathbb{R}^M$ be the feature representation generated by the feature extractor for the input instance $\mathbf{x}_j$ from the image patch (i.e., the bag) $X$. Then, the pooling operation will generate a bag representation $\mathbf{z}$ for $X$. There are different pooling operations used in the literature \cite{wang2018revisiting}, with the most common being the maximum operator:
\begin{equation}
    \mathbf{z} = \max_{j=1,...,K}\left\{\mathbf{h}_j\right\},
\end{equation}
the mean operator:
\begin{equation}
    \mathbf{z} = \frac{1}{K}\sum_{j=1}^K{\mathbf{h}_j},
\end{equation}
and the log-sum-exp (Mellomax) operator:
\begin{equation} \label{eq:4}
    \mathbf{z} = \frac{1}{r} \log{ \left( \frac{1}{K}\sum_{j=1}^K{\exp{r\mathbf{h}_j}} \right)},
\end{equation}
with $r \in (0,+\infty)$ being a meta-parameter, where $r\rightarrow0$ makes Equation (\ref{eq:4}) approximate the mean operator, whereas $r>>0$ makes Equation (\ref{eq:4})) a smooth approximation of the maximum operator. However, these MIL pooling operations have the disadvantage to be predefined and non-trainable. Ilse \textit{et al.} \cite{ilse2018attention} proposed to use instead a weighted average where the weights are computed by a trainable attention mechanism:
\begin{equation}
    \mathbf{z} = \sum_{j=1}^K{\alpha_j\mathbf{h}_j},
\end{equation}
where $\alpha_j$ can be computed by either an Attention layer:
\begin{equation}
    \alpha_j = \frac{\exp{\left\{ \mathbf{w}^\top \tanh{\left( \mathbf{Vh}_j^\top \right)}\right\}}}{\sum_{k=1}^K{\exp{\left\{ \mathbf{w}^\top \tanh{\left( \mathbf{Vh}_k^\top \right)}\right\}}}}
\end{equation}
or a Gated Attention layer:
\begin{equation}
    \alpha_j = \frac{\exp{\left\{ \mathbf{w}^\top \left[ \tanh{\left( \mathbf{Vh}_j^\top \right)} \odot \text{sigm}{\left( \mathbf{Uh}_j^\top \right)} \right] \right\}}}{\sum_{k=1}^K{\exp{\left\{ \mathbf{w}^\top \left[ \tanh{\left( \mathbf{Vh}_k^\top \right)} \odot \text{sigm}{\left( \mathbf{Uh}_k^\top \right)} \right] \right\}}}}
\end{equation}
where $\mathbf{w}\in \mathbb{R}^{L \times 1}$, $\mathbf{V}\in \mathbb{R}^{L \times M}$ and $\mathbf{U}\in \mathbb{R}^{L \times M}$ are parameters, $L$ is the number of hidden features adopted in the attention layers, $\odot$ is the element-wise multiplication, and $\text{tanh}$ and $\text{sigm}$ are the element-wise hyperbolic tangent and sigmoid activation functions, respectively.
\par
\begin{figure} [ht]
   \begin{center}
   \begin{tabular}{c} 
   \includegraphics[height=9cm]{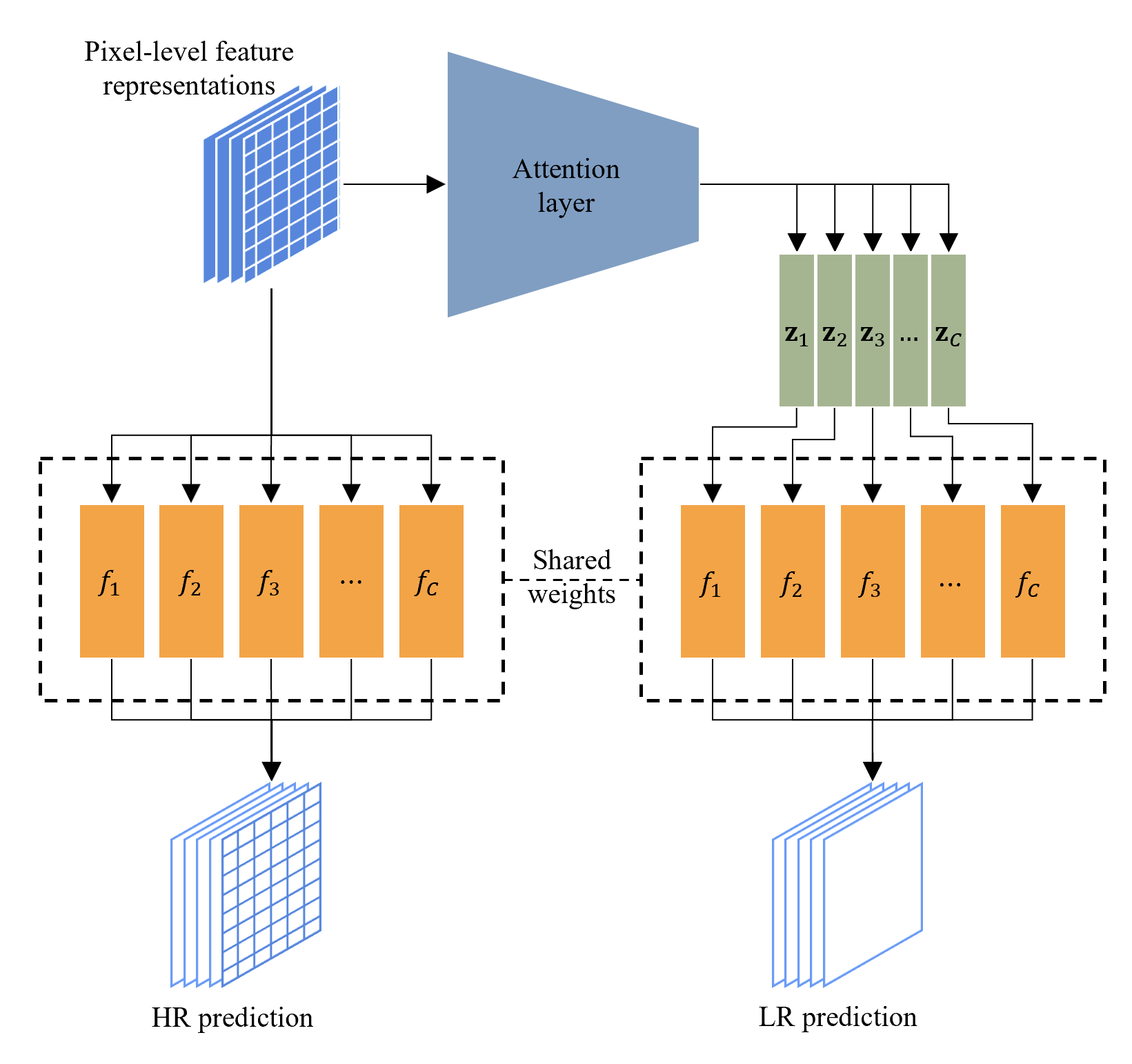}
   \end{tabular}
   \end{center}
   \caption[method] 
   { \label{fig:method_admil} 
Architecture of the proposed DMIL module. The attention layer generates the class bag representations that are classified with the same classifiers used to generate the HR predictions. 
}
   \end{figure}

Note that these attention-based strategies are all defined over binary classification tasks, and thus requires to be adapted to the multi-class scenario with $C$ classes. Standard approaches such as \textit{mean} and \textit{log-sum-exp} can instead be easily extended to the multi-class case. Here, we propose a straightforward adaptation of attention-based approaches by adopting an attention layer for each class. Hence, the bag of instances is used to produce $C$ different bag representations $\mathbf{z}_i, i=1,...,C$. Note that the instance representations and the bag representation all lie in the same feature space, hence a classifier can be used seamlessly on both. A binary linear classifier $f_i$ is used to compute a score for each class $\omega_i, i=1,...,C$. Then, class posterior probabilities can be estimated by applying the \textit{softmax} operator to the vector of bag class scores $f_i(\mathbf{z}_i),i=1,...,C$ or to the vector of pixel class scores $f_i(\mathbf{h}_j), \forall  i,j$.
Here, we also propose an alternative to the Gated Attention layer, which uses the GeLU as activation function in place of Tanh and Sigmoid, avoiding the saturating behaviour of these activations:
\begin{equation}
    \alpha_j = \frac{\exp{\left\{ \mathbf{w}^\top \left[ \text{GeLU}{\left( \mathbf{Vh}_j^\top \right)} \odot \text{GeLU}{\left( \mathbf{Uh}_j^\top \right)} \right] \right\}}}{\sum_{k=1}^K{\exp{\left\{ \mathbf{w}^\top \left[ \text{GeLU}{\left( \mathbf{Vh}_k^\top \right)} \odot \text{GeLU}{\left( \mathbf{Uh}_k^\top \right)} \right] \right\}}}}.
\end{equation}
Fig. \ref{fig:method_admil} shows the architecture of the proposed DMIL module.
\subsection{Training Strategy}
While MIL provide us a framework for training an HR land-cover classifier using LR land-cover maps, some considerations are necessary. Indeed, standard MIL is defined over binary classification tasks, thus it should be re-framed into a multi-class classification problem. The MIL assumption of Equation (\ref{eq:1}) tells us that it is sufficient that at least one pixel of the patch belongs to a certain class so that the patch can be labelled as that class. This implies that in a multi-class formulation, the MIL problem is also multi-label, as more than one class can be associated to an image patch at a time. However, there is only a single LR label associated to the given patch. Thus, the given LR labels provide only positive labels for each class, whereas the absence of positive labels cannot be interpreted as a negative label. To manage this, two approaches can be taken, based on two alternative assumptions:
\begin{itemize}
    \item \textit{Multi-class assumption}: The relationship between low- and high- resolution labels is different from the standard MIL assumption of Equation (\ref{eq:1}). A reasonable alternative is assuming that the LR label is representative of the majority of the HR land-cover labels, and Equation (\ref{eq:1}) can be rewritten as follows:
    \begin{equation}
        Y_i=\begin{cases}
            +1 & \sum_{j=1}^K{y_{ij}}>\sum_{j=1}^K{y_{ij}}, \quad \forall l \neq i \\
            0 & \text{otherwise}
        \end{cases},
    \end{equation}
    where $Y_i \in \{0,+1\}$ is the LR label associated to class $\omega_i$, and $y_{ij} \in \{0,+1\}$ is the true HR land-cover label of pixel $j$ for class $\omega_i$. Since only one bag label can be positive at a time, this assumption allows us to consider $Y_i=0$ as a negative label, and thus to solve the MIL problem with standard approaches, e.g., by minimizing the empirical risk on the LR labels, based on the categorical cross-entropy loss on the LR prediction.
    \item \textit{Multi-label assumption}: The relationship between low- and high- resolution labels expressed in Equation (\ref{eq:1}) is still valid. Here, the label definition is modified to assume three different meanings $Y_i =-1,0,+1$, which are negative, unlabelled, and positive, respectively. However, negative labels are not provided under the \textit{multi-label} assumption, i.e., $Y_i \in \{ 0,+1\}$, as the absence of a positive label cannot be considered as a negative label. Therefore, a PUL approach is necessary to properly define a training strategy for the MIL model.
\end{itemize}
Training the MIL model under the \textit{multi-class} case is straightforward, whereas PUL requires dedicated training strategies. Here, the nnPU\cite{kiryo2017positive} strategy is adopted. Let us consider the binary classification task. Let $p(X,Y)$ be the joint density distribution of tuples $(X,Y)$, $p_p(X)=p(X \mid Y=+1)$ and $p_n(X)=p(X \mid Y=-1)$ be the positive and negative class-conditional density, respectively, $\pi_p=P(Y=+1)$ be the positive class-prior probability, and $\pi_n=P(Y=-1)=1-\pi_p$ be the negative class-prior probability. We consider the training set as the union of two datasets $\mathcal{X}_p$ and $\mathcal{X}_u$ independently sampled from $p_p(X)$ and $p(X)$, respectively. Let $f: \mathbb{R}^n \rightarrow \mathbb{R}$ be an arbitrary decision function, and $\ell: \mathbb{R}\times \{ -1, +1 \} \rightarrow \mathbb{R}$ be a loss function. We can define the positive and negative risks as $R_p^+(f)=\mathbb{E}_{X\sim p_p}[\ell(f(X),+1)]$ and $R_n^-(f)=\mathbb{E}_{X\sim p_n}[\ell(f(X),-1)]$, respectively. This allows us to write the standard (positive-negative learning) risk of $f$ as a combination of $R_p^+(f)$ and $R_n^-(f)$ as follows:
\begin{equation}
    R_{pn}(f) = \pi_p R_p^+(f) + \pi_n R_n^-(f).
\end{equation}
In PUL, an unbiased estimator of the risk $R_{pn}$ can be computed given the knowledge of $\pi_p$ as follows:
\begin{equation}
    R_{pu}(f) = \pi_p R_p^+(f) - \pi_p R_p^-(f) +  R_u^-(f),
\end{equation}
where $R_p^-(f)=\mathbb{E}_{X\sim p_p}[\ell(f(X),-1)]$ and $R_u^-(f)=\mathbb{E}_{X\sim p}[\ell(f(X),-1)]$. Then, it is straightforward to define the empirical version of the risk $R_{pu}(f)$ based on the datasets $\mathcal{X}_p$ and $\mathcal{X}_u$. However, using this risk has shown overfitting problems due to the possibility that $R_u^-(f) - \pi_p R_p^-(f)<0$. Hence, a non-negative version was derived \cite{kiryo2017positive}:
\begin{equation}
    \tilde{R}_{pu}(f) = \pi_p R_p^+(f) + \max\{0,R_u^-(f) - \pi_p R_p^-(f)\}.
\end{equation}
Note that this is defined for binary classification. In this paper, we propose to use the nnPU risk separately and independently on each land-cover class in the LR predictions. Moreover, the nnPU requires the knowledge of the class-prior probabilities. In the MIL \textit{multi-label} settings, this means knowing the prior probability that at least one pixel in the patch belongs to a given land-cover class. In practice, they can be estimated with the availability of few reference data, and here we assume that these class-prior probabilities are known.
\par
The \textit{multi-class} and \textit{multi-label} assumptions allows us to define two different strategies for training a DMIL model. In the \textit{multi-class} case, we minimize the empirical risk over the training set $\mathcal{D}=\{(X^{(m)},\mathbf{Y}^{(m)})\}_{m=1}^M$ on the LR one-hot vector labels $\mathbf{Y}^{(m)}=[Y_1^{(m)},...,Y_i^{(m)},...,Y_C^{(m)}]^\top$, based on the categorical cross-entropy loss on the LR prediction $f_i(X^{(m)}), i=1,...,C$:
\begin{equation}
    \mathcal{R}_{MC}= \frac{1}{M} \sum_{m=1}^M{\left[ -\sum_{i=1}^C{Y_i^{(m)} \log{\left(\frac{\exp{f_i(X^{(m)})}}{\sum_{k=1}^C{\exp{f_k(X^{(m)})}}}\right)}} \right]}.
\end{equation}
Instead, the empirical risk minimized under the \textit{multi-label} assumption is defined as the average of the nnPU empirical risk $\tilde{R}_{pu}(f)$ of each class, computed separately. Let $\bm{\pi}$ be the vector containing the prior probabilities of the positive label of each class, i.e., $\pi_i = P(Y_i=+1)$, and $\mathbf{p}=\sum_{m=1}^M\mathbf{Y}^{(m)}$ be the vector containing the number of positive labelled instances of each class in the training dataset $\mathcal{D}$, i.e., $p_i = \sum_{m=1}^M{Y_i^{(m)}}$. The \textit{multi-label} empirical risk can be defined as follows:
\begin{multline}
    \mathcal{R}_{ML} = \frac{1}{C}\sum_{i=1}^C\Biggl\{ \frac{\pi_i}{p_i} \left[ \sum_{m=1}^MY_i^{(m)}\ell\left[f_i(X^{(m)}),+1\right]\right] + \\ + \max\left\{ 0, \frac{1}{M-p_i} \left[ \sum_{m=1}^M(1-Y_i^{(m)})\ell\left[f_i(X^{(m)}),-1\right]\right] -\frac{\pi_i}{p_i} \left[ \sum_{m=1}^MY_i^{(m)}\ell\left[f_i(X^{(m)}),-1\right]\right]\right\} \Biggl\},
\end{multline}
where the Sigmoid loss function $\ell [z,y]=1/(1+\exp{(yz)})$ is used.
\par
An important part of the proposed method is the choice of which assumption to use in the learning procedure. The optimal choice mainly depends on the legend definition, but other factors can be relevant, such as the accuracy of the LR map adopted for training. Indeed, a PUL strategy as nnPU may be more robust against label noise than a standard multi-class strategy. Moreover, simply choosing one assumption over the other may result in a suboptimal solution. Therefore, we propose for the general case a combination of the two assumptions. This is done by combining the two empirical risks as follows:
\begin{equation}
    \mathcal{R} = \beta \mathcal{R}_{MC} + (1-\beta) \mathcal{R}_{ML},
\end{equation}
where $\beta \in [0,1]$ is a meta-parameter tuning the contribution of each risk during the training. The optimal value for $\beta$ can be found through meta-parameter tuning with a validation set before training the final DMIL model.
\section{DATASET DESCRIPTION AND EXPERIMENTAL RESULTS}
\label{sec:results}
This section presents the dataset employed for the experimental analysis, the experimental setup used to assess the effectiveness of the proposed PU-MIL architecture, and the evaluation of the experimental results from both the quantitative and qualitative view points.
\subsection{Dataset Description}
In the considered experimental setup, the 2020 IEEE GRSS Data Fusion Contest (DFC2020) dataset \cite{dfc2020, robinson2021global} has been used. The aim of the contest was to encourage research in large-scale land-cover mapping using weakly supervised strategies for learning from satellite data available worldwide. The contest employed the SEN12MS  dataset\cite{sen12ms, isprs-annals-IV-2-W7-153-2019}, which consists of tuples of three types of data: Sentinel-1 SAR images, Sentinel-2 multispectral images, and MODIS-derived land-cover maps. Despite the data are presented with a ground sample distance (GSD) of $10m$, the Sentinel images inherently possess a resolution ranging between $10m$ and $60m$, whereas the native resolution of the MODIS-derived land-cover map stands at $500m$. Consequently, the challenge lies in developing robust models capable of accurately predicting HR land cover using LR annotations. To assess and test the models, the DFC2020 dataset provides semi-manually generated HR land-cover maps of scenes with undisclosed geographical information which are not included within the SEN12MS dataset. Each triplet in SEN12MS and each quadruplet in DFC2020 consists of 3D tensors representing a patch, each with a spatial extent of $256\times 256$ pixels. SEN12MS consists of $180 662$ globally sampled patches, whereas DFC2020 consists of $6114$ patches collected from seven globally distributed cities ($986$ of which are used for validation and $5128$ for testing). Each patch is associated to four MODIS-derived LR land-cover maps, each following a different classification scheme: IGBP, LCCS land cover, LCCS land use, and LCCS surface hydrology. The HR semi-manually generated HR land-cover reference maps follow a simplified IGBP classification scheme. Tab. \ref{tab:legend} shows the IGBP classification scheme and the corresponding simplified version used in the HR reference maps, as well as the class distributions of both the LR and HR land-cover maps, which are quite different and unbalanced. The LR maps show low percentages of the Shrubland, Wetlands and Barren land covers, while the HR maps do not include the Savanna class. Both LR and HR land-cover maps in the DFC2020 dataset lack the Snow/Ice land cover.
\begin{table}[ht]
\centering
\caption{IGBP and corresponding simplified IGBP classification schemes used in the DFC2020 dataset. The class distribution (\%) is also reported for the full DFC2020 dataset.}
\scriptsize
\label{tab:legend}
\begin{tblr}{
  row{1} = {m,l},
  column{1} = {c},
  column{3} = {c},
  column{6} = {c},
  column{7} = {c},
  cell{1}{1} = {l},
  cell{1}{6} = {l},
  cell{1}{7} = {l},
  cell{2}{3} = {r=5}{},
  cell{2}{4} = {r=5}{JapaneseLaurel},
  cell{2}{5} = {r=5}{},
  cell{2}{6} = {r=5}{},
  cell{2}{7} = {r=5}{},
  cell{7}{3} = {r=2}{},
  cell{7}{4} = {r=2}{Roti},
  cell{7}{5} = {r=2}{},
  cell{7}{6} = {r=2}{},
  cell{7}{7} = {r=2}{},
  cell{9}{3} = {r=2}{},
  cell{9}{4} = {r=2}{Broom},
  cell{9}{5} = {r=2}{},
  cell{9}{6} = {r=2}{},
  cell{9}{7} = {r=2}{},
  cell{11}{4} = {Lime},
  cell{12}{4} = {SpringGreen},
  cell{13}{3} = {r=2}{},
  cell{13}{4} = {r=2}{Crail},
  cell{13}{5} = {r=2}{},
  cell{13}{6} = {r=2}{},
  cell{13}{7} = {r=2}{},
  cell{15}{4} = {SilverChalice},
  cell{16}{4} = {Aquamarine},
  cell{17}{4} = {Milan},
  cell{18}{4} = {Blue},
  hline{1} = {1}{-}{},
  hline{1} = {2}{-}{},
  hline{2,7,9,11-13,15-18} = {-}{},
  hline{19} = {1}{1-7}{},
  hline{19} = {2}{1-7}{},
}
{IGBP\\Class\\Number} & IGBP Class Name & {Aggregated\\Class\\Number} &  & {Simplified\\Class Name} & {LR Class\\Distribution (\%)} & {HR Class\\Distribution (\%)}\\
1 & Evergreen Needleleaf Forest & 1 &  & Forest & 16.3 & 23.1 \\
2 & Evergreen Broadleaf Forest &  &  &  &  & \\
3 & Deciduous Needleleaf Forest &  &  &  &  & \\
4 & Deciduous Broadleaf Forest &  &  &  &  & \\
5 & Mixed Forest &  &  &  &  & \\
6 & Closed Shrublands & 2  &  & Shrubland  &  0.6 & 6.0 \\
7 & Open Shrublands &  &  &  &  & \\
8 & Woody Savannas & 3  &  & Savanna & 18.4 & — \\
9 & Savanna &  &  &  & \\
10 & Grasslands & 4 &  & Grassland & 11.1 & 11.6 \\
11 & Permanent Wetlands & 5 &  & Wetlands &  1.2 & 7.0 \\
12 & Croplands & 6  &  & Croplands & 17.8 & 17.0 \\
13 & Croplands / Natural Vegetation Mosaics &  &  &  &  & \\
14 & Urban and Built-up Lands & 7 &  & Urban / Built-up & 10.2 & 10.6 \\
15 & Permanent Snow and Ice & 8 &  & Snow / Ice & — & — \\
16 & Barren & 9 &  & Barren &  0.1 & 2.3 \\
17 & Water Bodies & 10 &  & Water & 24.4 & 22.0
\end{tblr}
\end{table}
\par
\subsection{Design of Experiments}
In the proposed work, only Sentinel-2 multispectral images were used for training, neglecting the availability of Sentinel-1 SAR images. The pre-processing step involved clipping the top-of-atmosphere multispectral images in the interval $[0,10^4]$, rescaling, and computing several spectral indices (e.g., NDVI, NDWI, NDMI and NDBI). Only $10-20m$ spectral band are considered for training the models. The DFC2020 provides us the LR land-cover maps upsampled at a GSD of $10m$. However, the DMIL models require the LR labels at native resolution. In the considered experimental setup, we downsample the LR maps by means of a majority vote filter with kernel size $64 \times 64$ pixels and strides of the same size, reducing the LR land-cover maps to a size of $4 \times 4$ pixels. Note that in this way the LR resolution labels are made even coarser (i.e., $640m$) than the native resolution of $500m$.
\par
The considered experimental setup is aligned to the Data Fusion Contest. The training of the DMIL models is performed using only the LR simplified IGBP land-cover maps. However, in order to select the optimal meta-parameters (e.g., learning rate, weight decay and the PU-DMIL $\beta$ parameter), a separated and independent HR validation set is adopted (i.e., DFC2020). For this reason, the proposed work cannot be directly compared with best-performing methods of the contest, as tuning the meta-parameters in this way during the contest would not have been possible.
\par
\begin{table}[ht]
\centering
\caption{Land-cover classes distribution in both LR and HR land-cover maps of the validation and test subsets of DFC2020.}
\small
\label{tab:dfc_dist}
\begin{tblr}{
  row{1} = {}{c,m},
  row{2} = {m,l},
  column{3} = {c},
  column{4} = {c},
  column{5} = {c},
  column{6} = {c},
  cell{2}{3} = {l},
  cell{2}{4} = {l},
  cell{2}{5} = {l},
  cell{2}{6} = {l},
  cell{1}{3} = {c=2}{},
  cell{1}{5} = {c=2}{},
  cell{3}{1} = {JapaneseLaurel},
  cell{4}{1} = {Roti},
  cell{5}{1} = {Broom},
  cell{6}{1} = {Lime},
  cell{7}{1} = {SpringGreen},
  cell{8}{1} = {Crail},
  cell{9}{1} = {SilverChalice},
  cell{10}{1} = {Milan},
  cell{11}{1} = {Blue},
  hline{1} = {1}{-}{},
  hline{1} = {2}{-}{},
  hline{3} = {-}{},
  hline{12} = {1}{1-6}{},
  hline{12} = {2}{1-6}{},
}
  &  & DFC2020 Validation & & DFC2020 Test & \\
  & {Simplified\\Class Name} & {LR Class\\Distribution (\%)} & {HR Class\\Distribution (\%)} & {LR Class\\Distribution (\%)} & {HR Class\\Distribution (\%)}\\
  & Forest & 5.0 & 9.1 & 18.5 & 25.6 \\
  & Shrubland & 1.0 & 5.2 & 0.5 & 5.9 \\
  & Savanna & 36.8 & — & 14.8 & — \\
  & Grassland & 6.1 & 11.8 & 12.0 & 10.1 \\
  & Wetlands &  0.5 & 17.4 & 1.3 & 2.1 \\
  & Croplands & 10.0 & 13.0 & 19.3 & 19.5 \\
  & Urban / Built-up & 6.0 & 5.4 & 11.0 & 10.7 \\
  & Barren &  0.1 & 2.9 & 0.05 & 2.6 \\
  & Water & 34.5 & 35.0 & 22.5 & 23.3
\end{tblr}
\end{table}
Despite only DFC2020 is adopted in the experiments, the validation and test subsets are quite different, which is evident when looking at the class distributions reported in Tab. \ref{tab:dfc_dist}. Thus, we considered two training scenarios: i) the meta-parameters tuning is performed with the DFC2020 validation subset, and the full DFC2020 test subset is used for testing the selected models; and ii) only the DFC2020 test subset is considered, and the meta-parameters tuning is performed using $20\%$ of the subset,  while the remaining $80\%$ is used for testing the selected models. This allows us to consider the effects of the domain shift on the model selection procedure. Note that the same imagery is used both for the training and the evaluation of the models, with the difference that during training only LR land-cover maps are considered, whereas the evaluation of the model includes the HR land-cover maps. Moreover, the training set for meta-parameters tuning is always different from the training set adopted for testing.
\par
An evident difference between the LR and HR land-cover labels is the absence of the Savanna class among the HR land-cover labels. To address this issue, the same approach adopted by the winning team of the second track of the contest is taken\cite{robinson2021global}: the LR Savanna labels are considered as Wetlands when the full $256 \times 256$ pixels scene contains the Water class, otherwise they are considered as Grassland. Moreover, both the validation and test subsets show class imbalance. The most affected classes are the Barren and Shrubland classes ($\leq 1\%$), followed by the Wetlands class ($<2\%$). To mitigate this problem, the following procedure is considered. First, each of the $256 \times 256$ pixels patch is split into $16$ non-overlapping $64 \times 64$ pixels patches. The benefit of doing this is twofold: i) generating an increased diversity of land cover in the training batches, and ii) assigning each patch with a single LR label. Then, during training, image patches are sampled uniformly from each LR land-cover class. Moreover, random flip and $90$-degree rotations are applied every time a patch is sampled.
\par
The proposed PU-DMIL strategy requires the knowledge of “bag” class-prior distribution of the HR land-cover labels inside a patch. Here, we assume that this information is given, and class priors are computed directly from the DFC2020 validation and test subsets using Equation (\ref{eq:1}), also considering the influence of the oversampling. 
The models are trained using the Adam optimizer and a batch size of $64$ image patches. The default values are used for all the optimizer's parameters except for the learning rate and weight decay, which are treated as meta-parameters and tuned using the validation set. The meta-parameters considered in the experiments vary on the basis of the adopted model. Apart from the learning rate and weight decay, the experiment-dependent meta-parameters are the $\beta$ parameter in the empirical risk of the proposed PU-DMIL model and the $r$ parameter of the \textit{log-sum-exp} DMIL module. The meta-parameters tuning applies the Tree-structured Parzen Estimator (TPE)\cite{NIPS2011_86e8f7ab} algorithm to efficiently explore the meta-parameter space to optimize the average accuracy (i.e., mean class-wise producer's accuracy) after five training epochs. For each meta-parameter, TPE requires a probability distribution from which the meta-parameter values are initially sampled. Learning rate, weight decay and $r$ are sampled from a log-uniform distribution in the range $\left[ 10^{-5}, 10^{-2} \right]$, $\left[ 10^{-6}, 10^{-2} \right]$ and $\left[ 10^{-4}, 10^{5} \right]$, respectively, whereas $\beta$ is sampled uniformly from the interval $\left[ 0,1 \right]$. After tuning, the selected meta-parameters are used to train the model on the imagery and LR reference maps of the test set, which is the full or $80\%$ version of the DFC2020 test subset depending on the data used for validation. The selected models are trained five times for ten epochs with different randomly initialized weights, and the trained models with the median AA score on the test set are retained for the intercomparison with the other architectures. The metrics considered are class-wise producer's accuracies, Average Accuracy (AA), and the mean Intersection over Union (mIoU), which are well suited for semantic segmentation tasks such as land-cover mapping.
\par
The models considered in the experiments are the standard model (Std) as shown in Fig. \ref{fig:method}(a) and the architectures adopting the DMIL modules considered in Section \ref{sec:mil_method}, i.e., Mean, log-sum-exp (LSE), Attention (Attn), Gated Attention (GAttn) and the proposed alternative to Gated Attention (Prop).
\subsection{Experimental Results}
Tab. \ref{tab:meta_params} shows the final meta-parameters selected by the TPE algorithm for all the considered models, using both the DFC2020 validation and test ($20\%$) subsets. One can notice that the selected meta-parameters are quite different based on which dataset is used for meta-parameters tuning. Specifically, when examining experiments conducted on the same dataset, the learning rates consistently exhibit a similar order of magnitude. However, a contrasting order of magnitude becomes apparent when comparing experiments conducted on distinct subsets.
\begin{table}[ht]
\centering
\caption{Final meta-parameters selected by the TPE algorithm for each experiment performed. The table also shows the difference between using the validation subset \textit{vs.} using $20\%$ of the test subset.}
\small
\label{tab:meta_params}
\begin{tblr}{
  row{1} = {m,l},
  row{2} = {m,l},
  column{2} = {m,r},
  column{3} = {m,c},
  column{4} = {m,c},
  column{5} = {m,c},
  column{6} = {m,c},
  cell{2}{1} = {r=6}{},
  cell{8}{1} = {r=6}{},
  hline{1} = {1}{-}{},
  hline{1} = {2}{-}{},
  hline{2,8} = {-}{},
  hline{14} = {1}{1-7}{},
  hline{14} = {2}{1-7}{},
}
                       & Model & Learning rate        & Weight Decay         & $\beta$ & $r$                  & AA (\%)\\
{DFC2020\\Validation}  & Std   & $9.5 \times 10^{-5}$ & $2.1 \times 10^{-3}$ & —       & —                    & 64.4   \\
                       & Mean  & $2.2 \times 10^{-5}$ & $1.8 \times 10^{-3}$ & $0.80$  & —                    & 64.5   \\
                       & LSE   & $2.4 \times 10^{-5}$ & $2.0 \times 10^{-6}$ & $0.72$  & $1.1 \times 10^{-3}$ & 64.5   \\
                       & Attn  & $3.0 \times 10^{-5}$ & $4.6 \times 10^{-3}$ & $0.09$  & —                    & 65.1   \\
                       & GAttn & $1.2 \times 10^{-5}$ & $3.7 \times 10^{-4}$ & $0.49$  & —                    & 65.2   \\
                       & Prop  & $4.0 \times 10^{-3}$ & $3.3 \times 10^{-5}$ & $0.48$  & —                    & 65.5   \\
{DFC2020\\Test $20\%$} & Std   & $2.4 \times 10^{-4}$ & $2.4 \times 10^{-6}$ & —       & —                    & 56.0   \\
                       & Mean  & $8.4 \times 10^{-4}$ & $5.6 \times 10^{-4}$ & $0.99$  & —                    & 58.1   \\
                       & LSE   & $7.7 \times 10^{-4}$ & $1.3 \times 10^{-3}$ & $0.81$  & $3.5 \times 10^{-4}$ & 58.4   \\
                       & Attn  & $4.1 \times 10^{-4}$ & $1.3 \times 10^{-6}$ & $0.77$  & —                    & 58.7   \\
                       & GAttn & $1.7 \times 10^{-4}$ & $5.7 \times 10^{-4}$ & $0.91$  & —                    & 58.2   \\
                       & Prop  & $3.7 \times 10^{-4}$ & $1.5 \times 10^{-3}$ & $0.74$  & —                    & 58.0   \\
\end{tblr}
\end{table}
\par
When considering the meta-parameters of the standard MIL strategies, such as Mean and LSE, we observe that the choice of the $\beta$ parameter is biased towards the \textit{multi-class} assumption. This can be imputed to the Mean and LSE models being unable to dynamically select different pixels for each class. Indeed, the Mean model is governed by the “average” instance, and thus a majority voting assumption such as the \textit{multi-class} assumption is better suited for this model. The LSE model is a generalized module that includes the \textit{mean} and \textit{max} MIL pooling operations as extreme cases of its parameter $r$. In the meta-parameters tuning phase, the best value for $r$ has shown to be close to $0$, which leads the LSE model to approximate the behaviour of the Mean model. Therefore, a similar argument to the one of the Mean model can be made. Moreover, larger values of $r$ lead to approximating the MIL pooling operation, thus focusing the attention on few pixels in each image patch, resulting in worse performance for HR land-cover mapping.
\par
When considering the meta-parameters of the attention-based PU-DMIL models, the $\beta$ parameter values showed interesting results. In the case of meta-parameters tuning with the DFC2020 validation subset, the best performing values are different from the Mean and LSE ones. In particular, the best value of $\beta$ for Attn is strongly biased toward the PUL strategy. Instead, the GAttn and Prop models found values for $\beta$ that equally weight the \textit{multi-class} and \textit{multi-label} risks. Overall, the attention-based PU-DMIL models were able to achieve the best AAs during meta-parameters tuning. However, one should remember that these scores are not general as they are biased by the meta-parameters tuning. In the case of meta-parameters tuning with $20\%$ of the DFC2020 test subset, the best performing values of $\beta$ leans towards the \textit{multi-class} risk. Among the attention-based PU-DMIL models, only the $\beta$ value for GAttn is strongly biased towards the \textit{multi-class} risk. Generally, there are large differences in the selected meta-parameters depending on the validation set used. This result, together with the differences in the learning rates, confirms that a domain shift is indeed present between the DFC2020 validation and test subsets.
\par
\begin{table}
\centering
\caption{Classification performance of the selected models on the DFC2020 test subset. The selected models are trained five times with randomly initialized weights, and the results of the model with median AA score are reported. The full test subset is used for models selected using the DFC2020 validation subset, whereas 80\% of the test subset is used for testing the models selected with the other 20\% of the test subset. Class-wise accuracy (i.e., producer's accuracy) are reported, as well as the AA and the mIoU. Best results are highlighted in gray.}
\footnotesize
\label{tab:results}
\begin{tblr}{
  cell{1}{3} = {c=6}{},
  cell{1}{9} = {c=6}{},
  cell{11}{1} = {c=2}{},
  cell{12}{1} = {c=2}{},
  row{1} = {m,c},
  row{2} = {m,l},
  column{1} = {m,r},
  column{3} = {m,c},
  column{4} = {m,c},
  column{5} = {m,c},
  column{6} = {m,c},
  column{7} = {m,c},
  column{8} = {m,c},
  column{9} = {m,c},
  column{10} = {m,c},
  column{11} = {m,c},
  column{12} = {m,c},
  column{13} = {m,c},
  column{14} = {m,c},
  cell{3}{2} = {JapaneseLaurel},
  cell{4}{2} = {Roti},
  cell{5}{2} = {Lime},
  cell{6}{2} = {SpringGreen},
  cell{7}{2} = {Crail},
  cell{8}{2} = {SilverChalice},
  cell{9}{2} = {Milan},
  cell{10}{2} = {Blue},
  cell{3}{6} = {Alto},
  cell{4}{3} = {Alto},
  cell{5}{7} = {Alto},
  cell{6}{3} = {Alto},
  cell{7}{8} = {Alto},
  cell{8}{3} = {Alto},
  cell{9}{7} = {Alto},
  cell{10}{8} = {Alto},
  cell{11}{8} = {Alto},
  cell{12}{8} = {Alto},
  cell{3}{12} = {Alto},
  cell{4}{12} = {Alto},
  cell{5}{14} = {Alto},
  cell{6}{10} = {Alto},
  cell{7}{11} = {Alto},
  cell{8}{9} = {Alto},
  cell{9}{11} = {Alto},
  cell{10}{14} = {Alto},
  cell{11}{14} = {Alto},
  cell{12}{12} = {Alto},
  vline{3,9} = {-}{},
  hline{1} = {1}{-}{},
  hline{1} = {2}{-}{},
  hline{3} = {-}{},
  hline{11} = {1}{-}{},
  hline{11} = {2}{-}{},
  hline{13} = {1}{1-14}{},
  hline{13} = {2}{1-14}{},
}
                & & DFC2020 Full Test Subset & & & & & & DFC2020 $80\%$ Test Subset & & & & & \\
Class Name      & & Std    & Mean    & LSE    & Attn    & GAttn    & Prop    & Std    & Mean    & LSE    & Attn    & GAttn    & Prop    \\
Forest          & & 0.74   & 0.75    & 0.77   & 0.80    & 0.74     & 0.74    & 0.67   & 0.77    & 0.75   & 0.83    & 0.80     & 0.80    \\
Shrubland       & & 0.20   & 0.18    & 0.19   & 0.19    & 0.16     & 0.19    & 0.20   & 0.21    & 0.18   & 0.27    & 0.20     & 0.21    \\
Grassland       & & 0.47   & 0.52    & 0.51   & 0.29    & 0.60     & 0.54    & 0.58   & 0.44    & 0.51   & 0.50    & 0.53     & 0.64    \\
Wetlands        & & 0.55   & 0.53    & 0.49   & 0.37    & 0.53     & 0.53    & 0.58   & 0.63    & 0.55   & 0.51    & 0.60     & 0.53    \\
Croplands       & & 0.65   & 0.68    & 0.68   & 0.66    & 0.68     & 0.73    & 0.58   & 0.72    & 0.75   & 0.62    & 0.68     & 0.61    \\
Built-up        & & 0.79   & 0.72    & 0.76   & 0.75    & 0.77     & 0.77    & 0.82   & 0.74    & 0.70   & 0.77    & 0.76     & 0.77    \\
Barren          & & 0.08   & 0.08    & 0.05   & 0.01    & 0.09     & 0.00    & 0.07   & 0.06    & 0.10   & 0.03    & 0.01     & 0.00    \\
Water           & & 0.92   & 0.94    & 0.93   & 0.96    & 0.92     & 1.00    & 0.90   & 0.92    & 0.94   & 0.96    & 0.95     & 0.98    \\
AA (\%)         & & 54.8   & 55.0    & 54.8   & 50.3    & 56.1     & 56.3    & 55.0   & 56.1    & 56.1   & 56.3    & 56.7     & 56.9    \\
mIoU (\%)       & & 40.9   & 41.6    & 41.8   & 37.8    & 42.2     & 42.7    & 40.0   & 42.3    & 42.7   & 43.2    & 43.1     & 42.9    \\ 
\end{tblr}
\end{table}
Tab. \ref{tab:results} shows the performance of the selected models when trained on the test set. In general, models whose meta-parameters are tuned on 20\% of the DFC2020 test subset perform better than models whose meta-parameters are tuned on the DFC2020 validation subset. This is expected given the domain shift between the two subsets of DFC2020. In both cases, PU-DMIL models always outperform the standard model in terms of both AA and mIoU.
The results also shows that the best performing architecture in terms of AA is the proposed alternative to the Gated Attention DMIL model (i.e., Prop), especially in the case of the meta-parameters tuned using 20\% of the DFC2020 test subset. However, Prop reports the worst performance on the Barren class, showing to be particularly sensitive to class imbalance, as the Barren class was strongly underrepresented in the LR reference data. Despite this, the AA obtained in this case is the best ever achieved in all the experiments, with a competitive score for the mIoU. The results of the Attn model are of particular interest, as its performance with the meta-parameters tuned using the DFC2020 validation subset is the worst in the same category, whereas its performance with the meta-parameters tuned using 20\% of the DFC2020 test subset is among the top-performing ones. In particular, Attn showed the best mIoU and the best performances on the Forest and Shrubland classes.
\par
\begin{figure}[t]
    \centering
    \subfigure[]{\includegraphics[width=0.18\textwidth]{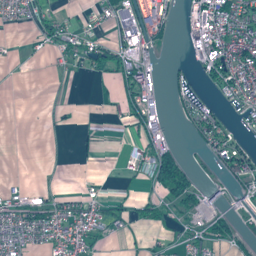}} 
    \subfigure[]{\includegraphics[width=0.18\textwidth]{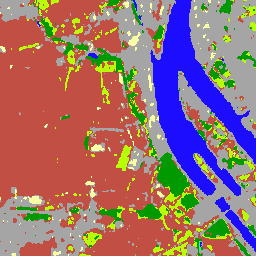}} 
    \subfigure[]{\includegraphics[width=0.18\textwidth]{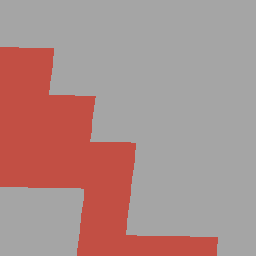}}
    \subfigure[]{\includegraphics[width=0.18\textwidth]{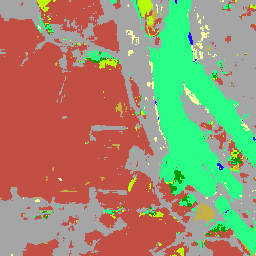}}
    \subfigure[]{\includegraphics[width=0.18\textwidth]{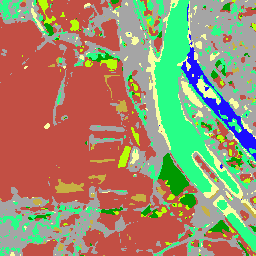}}
    \subfigure[]{\includegraphics[width=0.18\textwidth]{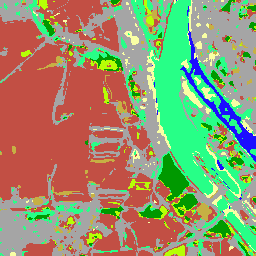}}
    \subfigure[]{\includegraphics[width=0.18\textwidth]{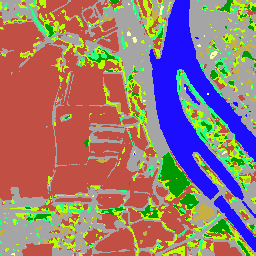}}
    \subfigure[]{\includegraphics[width=0.18\textwidth]{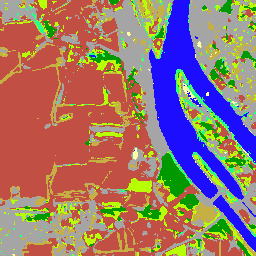}}
    \subfigure[]{\includegraphics[width=0.18\textwidth]{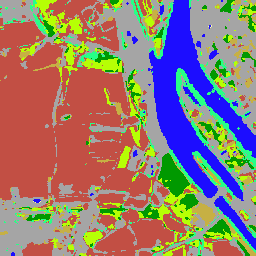}}
    \caption{Qualitative example of the classification maps obtained by the considered models: (a) Sentinel-2 RGB image, (b) Reference HR land-cover map, (c) Reference LR land-cover map used for training. Classification maps obtained by: (d) the standard model, (e) the Mean PU-DMIL model, (f) the Log-Sum-Exp PU-DMIL model, (g) the Attention PU-DMIL model, (h) Gated Attention PU-DMIL model, (i) the proposed alternative to the Gated Attention PU-DMIL model.}
    \label{fig:qualitative1}
\end{figure}
\begin{figure}[t]
    \centering
    \subfigure[]{\includegraphics[width=0.18\textwidth]{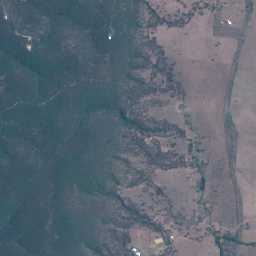}} 
    \subfigure[]{\includegraphics[width=0.18\textwidth]{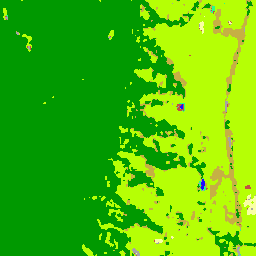}} 
    \subfigure[]{\includegraphics[width=0.18\textwidth]{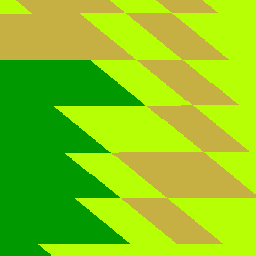}}
    \subfigure[]{\includegraphics[width=0.18\textwidth]{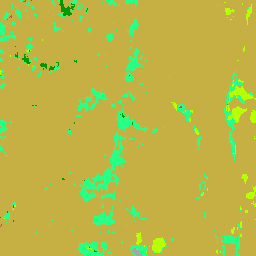}}
    \subfigure[]{\includegraphics[width=0.18\textwidth]{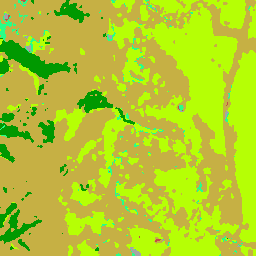}}
    \subfigure[]{\includegraphics[width=0.18\textwidth]{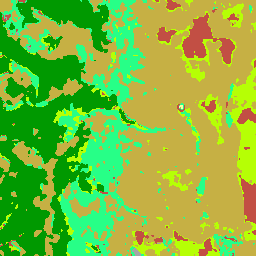}}
    \subfigure[]{\includegraphics[width=0.18\textwidth]{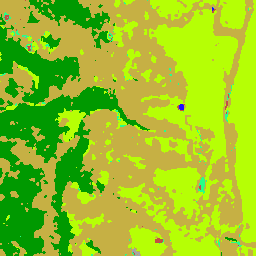}}
    \subfigure[]{\includegraphics[width=0.18\textwidth]{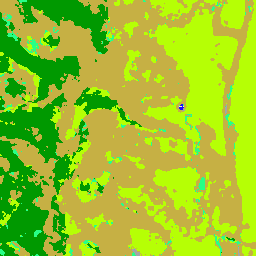}}
    \subfigure[]{\includegraphics[width=0.18\textwidth]{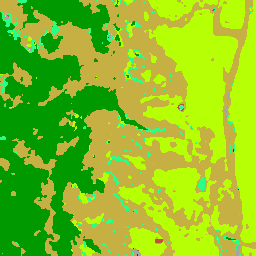}}
    \caption{Qualitative example of the classification maps obtained by the considered models: (a) Sentinel-2 RGB image, (b) Reference HR land-cover map, (c) Reference LR land-cover map used for training. Classification maps obtained by: (d) the standard model, (e) the Mean PU-DMIL model, (f) the Log-Sum-Exp PU-DMIL model, (g) the Attention PU-DMIL model, (h) Gated Attention PU-DMIL model, (i) the proposed alternative to the Gated Attention PU-DMIL model.}
    \label{fig:qualitative2}
\end{figure}

Fig. \ref{fig:qualitative1} and Fig. \ref{fig:qualitative2} show the qualitative results of the different methods with meta-parameters tuned on 20\% of the DFC2020 test subset. The images show how the trained attention-based DMIL models outperform both the baseline model and the standard DMIL models. In particular, the water class is better represented by the proposed PU-DMIL models, especially on rivers, which are rarely represented at $500m$ resolution. Moreover, the proposed attention-based PU-DMIL models are able to better capture the spatial patterns, as shown in Fig. \ref{fig:qualitative2}.

\section{CONCLUSION}
\label{sec:conclusion}
In this paper, we have introduced a framework for training DNNs for land-cover mapping using LR reference data, where each LR label is associated to an HR satellite image patch. Inspired by MIL, we have proposed DMIL architectures that can predict both LR and HR land-cover labels. Despite they are supervised with LR data, the DMIL model exploits a MIL pooling operator to link the LR reference map to the HR land-cover labels, allowing the model to learn the HR labels as by-product. The training of the DMIL architectures is performed considering two alternative assumptions. With the first assumption, the task at hand is the prediction of the dominant HR label in the patch. Instead, with the second one, the task is to predict all the land covers that are present in the given patch. However, this is an ill-posed multi-label problem where only positive labels are given for training. Hence, we proposed the combination of DMIL with a PUL strategy, which is able to approximate a fully supervised learning scenario thanks to the knowledge of the prior distribution of the land covers. Then, these two tasks are combined to take advantage from the benefits of both. Experimental results on the IEEE GRSS Data Fusion Contest 2020 show the effectiveness of the proposed strategy compared to standard techniques. The proposed method is promising, and several aspects can be considered for future improvements, such as the development of strategies for either handling the domain shift inherent to global land-cover mapping or strategies for better handling class imbalance. Moreover, we plan to integrate this work in a more generalized framework for weak supervision, where not only inexact supervision (which is addressed in this work) is considered, but also inaccurate supervision is considered, e.g., by exploiting label-noise robust loss functions or the partial knowledge of the label noise process\cite{9477628}.

\acknowledgments 
 
This research was supported by the European Space Agency (ESA) within the project CCI+ HRLC—Climate Change Initiative Extension (CCI+), Phase 1: New Essential Climate Variables (NEW ECVS) High Resolution Land Cover ECV (HR\_LandCover\_cci)

\bibliography{report} 
\bibliographystyle{spiebib} 

\end{document}